\documentclass[sn-mathphys]{sn-jnl}

\jyear{2021}%

\theoremstyle{thmstyleone}
\theoremstyle{thmstyletwo}%

\theoremstyle{thmstylethree}%

\begin{document}

\title{Semantic Feature Integration network for Fine-grained Visual Classification}


\author[1]{\fnm{Hui} \sur{Wang}}\email{6213113108@stu.jiangnan.edu.cn}

\author*[1]{ \fnm{Yueyang}  \sur{Li} }\email{lyueyang@jiangnan.edu.cn}

\author[2]{\fnm{Haichi} \sur{Luo}}\email{luohaichi@jiangnan.edu.cn}

\affil*[1]{\orgdiv{Jiangsu Provincial Engineering Laboratory of Pattern Recognition and Computational Intelligence}, \orgname{Jiangnan University}, \orgaddress{\street{1800 Lihu Avenue}, \city{Wuxi}, \postcode{214000}, \state{Jiangsu}, \country{China}}}

\affil[2]{\orgdiv{College of Internet of Things Engineering}, \orgname{Jiangnan University}, \orgaddress{\street{1800 Lihu Avenue}, \city{Wuxi}, \postcode{214000}, \state{Jiangsu}, \country{China}}}


\abstract{Fine-Grained Visual Classification (FGVC) is known as a challenging task due to subtle differences among subordinate categories. Many current FGVC approaches focus on identifying and locating discriminative regions by using the attention mechanism, but neglect the presence of unnecessary features that hinder the understanding of object structure. These unnecessary features, including 1) ambiguous parts resulting from the visual similarity in object appearances and 2) noninformative parts (e.g., background noise), can have a significant adverse impact on classification results. In this paper, we propose the Semantic Feature Integration network (SFI-Net) to address the above difficulties. By eliminating unnecessary features and reconstructing the semantic relations among discriminative features, our SFI-Net has achieved satisfying performance. The network consists of two modules: 1) the multi-level feature filter (MFF) module is proposed to remove unnecessary features with different receptive field, and then concatenate the preserved features on pixel level for subsequent disposal; 2) the semantic information reconstitution (SIR) module is presented to further establish semantic relations among discriminative features obtained from the MFF module. These two modules are carefully designed to be light-weighted and can be trained end-to-end in a weakly-supervised way. Extensive experiments on four challenging fine-grained benchmarks demonstrate that our proposed SFI-Net achieves the state-of-the-arts performance. Especially, the classification accuracy of our model on CUB-200-2011 and Stanford Dogs reaches 92.64\% and 93.03\%, respectively.}

\keywords{Fine-grained image classification, Semantic relation learning, Attention mechanism}



\maketitle

\begin{figure}[h]%
\centering
\includegraphics[width=0.99\textwidth]{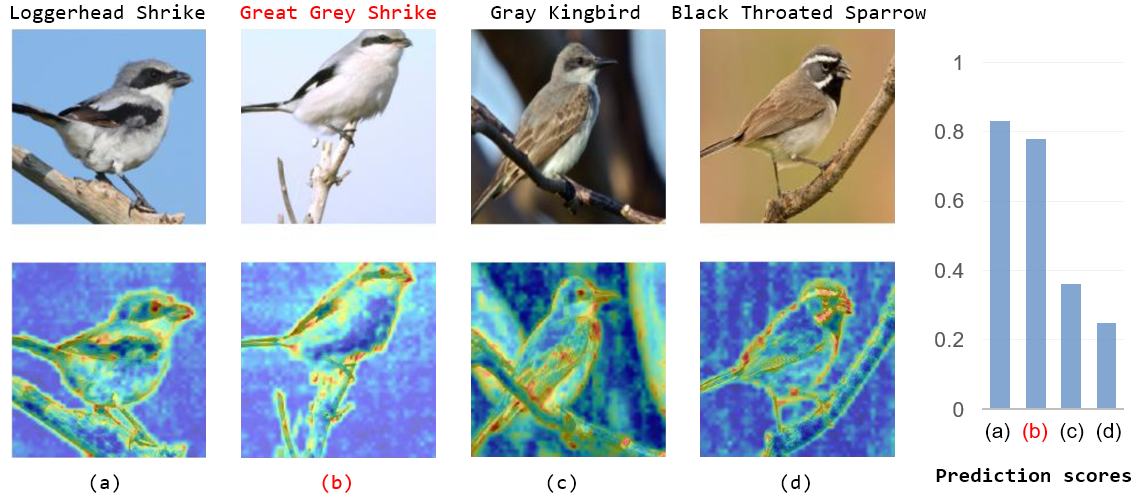}
\caption{Attention visualization of four misclassified bird images and their respective prediction scores. The left part shows a set of misclassified images and the feature maps extracted from the backbone network Swin transformer \cite{liu2021swin}. The bird specie marked in red represents the ground truth of prediction. The histogram visualized in the right part represents confident score of each class obtained from the softmax layer. The above visualizations indicate that although the baseline model can extract satisfying overall features of object, similar features of subcategories and the background noise can not be easliy averted. These unnecessary information results in misclassifications and thereby affects model's performance. The above images are from CUB-200-2011 dataset and the feature map visualization is generated through gradient-weighted class activation mapping \cite{selvaraju2017grad} technology.}\label{fig1}
\end{figure}
\section{Introduction}\label{sec1}
Fine-grained visual classification (FGVC) aims at identifying the sub-category of an image from the same superior class (e.g., identifying different types of birds). Different from general image classification task, FGVC is more challenging due to small inter-class variance and large intra-class variance. For one thing, sub-categories in FGVC share the striking resemblance. For example, "Loggerhead Shrike" and "Great Gray Shrike" have similar appearances and only differ in feathers' texture. As shown in Fig.~\ref{fig1}, though the overall features are captured precisely, the general classification model (e.g., Swin-T) is not capable to tell the difference between them, which results in a wrong prediction. For another, objects in the FGVC task often present different poses and perspectives though they belong to the same sub-category. Thus, many methods have been proposed to meet these challenges.

Early works are designed to extract discriminative features with the help of manual information, such as part locations and object bounding boxes \cite{zhang2014part,wei2016mask,branson2014bird,berg2013poof,huang2016part}. However, these approaches require expensive human annotations and can not be easily generalized to similar tasks. Aiming at solving this issue, scholars begin exploring weakly-supervised methods \cite{bib13,lin2015bilinear,zhuang2020learning,li2020attribute,du2020fine,liu2021subtler,behera2021context,fu2017look,ding2019selective,chen2022fine,liu2020filtration,hanselmann2020elope}, which only require image category labels for training. For example, attention mechanism is widely used to locate discriminative regions \cite{liu2021subtler,behera2021context,fu2017look,ding2019selective}. By establishing Encoder-Decoder framework on spatial or channel dimension, these methods can compare features and target labels for better localization. These located regions are thought as discriminative regions and will be assigned with more attention. Another class of methods \cite{hanselmann2020elope,chen2022fine,liu2020filtration} focus on eliminating noninformative parts, which promotes the importance of located discriminative image regions. However, as we analysed in Fig.~\ref{fig1}, the above methods generally ignore that these selected discriminative regions can include ambiguous parts (similar features that the top-k classes share) along with salient background noise (e.g. twig, flowers and water ripples). These unnecessary regions can cause misclassification due to irrelevant and confusing inormation.

In this paper, we propose the Semantic feature Integration network (SFI-Net) to address the aforementioned problems. To this end, we propose a novel module called multi-level feature filter (MFF) to remove both ambiguous parts and background noise based on the output features produced by the backbone network. This MFF module exploits pixel-level features with varying receptive field sizes to identify and eliminate unnecessary regions that may make the model puzzled, working as a feature selector to generate valid information. Although features obtained from MFF are purified, they are spatially scattered and lack euclidean structure. As a result, the semantic information reconstitution (SIR) module is introduced to process these valid, yet spatially irregular information by utilizing graph neural networks and the self-attention mechanism. The SIR module can enhance our model to mine spatial clues among these preserved discriminative features and rebuild their semantic structure.

The main contributions of our work are summarized as follows:

(1) We propose the semantic feature integration network termed SFI-Net specializing in fine-grained classification task. It includes two key components that help to effectively retain and reconstruct useful information.

(2) In order to retain valid information for fine-grained classification task, we design the multi-level feature filter (MFF) module to filter out ambiguous areas and background noise on pixel level. Then the semantic information reconstitution (SIR) module is introduced to integrate the preserved features for better accuracy.

(3) We validate the superiority of our SFI-Net through substantial experiments on four datasets, including CUB-200-2011 \cite{wah2011caltech},  Stanford Dogs \cite{khosla2011novel}, Stanford Cars \cite{krause20133d} and FGVC-aircraft \cite{maji2013fine}. The results demonstrate that our model achieves the state-of-the-art performance.

The rest of this paper is organized as follows. Sec.~\ref{sec2} reviews related works on fine-grained visual classification (FGVC) and the attention mechanism applied in FGVC task. Sec.~\ref{sec3} explains the details of the proposed SFI-Net. In Sec.~\ref{sec4}, the experimental results and a thorough analysis of the ablation study are presented. At last, Sec.~\ref{sec5} concludes this paper.



\section{Related work}\label{sec2}

In this part, we mainly introduce and discuss the previous work of fine-grained visual classification and related attention mechanism.

\subsection{Fine-grained visual classification}\label{subsec2}

Fine-grained visual classification aims to recognize objects from similar subordinate categories, which is a challenging task due to the subtle inter-class and large intra-class variation. In the meantime, discriminative parts for fine-grained classification tend to exist in smaller areas of the image compared with traditional image categorization. Most of the previous methods focus on discriminative regions discovery and feature extraction to solve the above problems\cite{zhang2014part,wei2016mask,branson2014bird,lin2015bilinear,zhuang2020learning,huang2021snapmix,zhang2018adversarial,hanselmann2020elope}. Early ones \cite{zhang2014part,branson2014bird,wei2016mask} rely on part-level annotations to guide the training of their models. For example, Part-based R-CNN \cite{zhang2014part} localize informative parts under a geometric prior. Branson et al. \cite{branson2014bird} utilize the pose-normalized representation to extract image features. However, these methods are impractical for further research due to expensive manual annotations.

 In order to solve this problem, researches have been conducted on the weakly-supervised methods that require only image-level annotations \cite{lin2015bilinear,zhuang2020learning,huang2021snapmix,zhang2018adversarial,hanselmann2020elope}. B-CNN \cite{lin2015bilinear} extracts features through two different CNN stream, then fuses them with matrix outer product. Zhuang et al. \cite{zhuang2020learning} construct a pairwise interaction network based on human intuition to capture contrastive clues. Simultaneously, many of these works focus on removing background noise to get effective object features. Huang et al. \cite{huang2021snapmix} propose SnapMix to exploit class activation map for fine-grained data augmentation. Zhang et al. \cite{zhang2018adversarial} propose an Adversarial Complementary Learning (ACoL) mechanism to localize objects of semantic interest. In \cite{hanselmann2020elope}, Hanselmann et al. train a bounding box detector to generate bounding boxes of the objects for image cropping. Although these above methods can alleviate the impact of background noise to some extent and have achieved satisfying results, these works generally locate discriminative regions on coarse image level, which may cause the retention of background noise and incorrect elimination of object parts. Our SFI-Net processes features all on pixel level, thereby avoids the above problems caused by coarse image cropping.

\subsection{Attention mechanism}\label{subsec2}

The attention mechanism aims to find correlations 
in original data and highlight some of its important characteristics, which fits the core needs of fine-grained classification (i.e. localizing discriminative regions). As a result, various of attention mechanisms have been incorporated into recent FGVC works \cite{fu2017look,ding2019selective,behera2021context,ding2021ap}. Fu et al. \cite{fu2017look} propose the recurrent attention network that learns region attention and region-based feature representation. Ding et al. \cite{ding2019selective} propose a selective sparse sampling framework to mine diverse and fine-grained details. Behara et al. \cite{behera2021context} design the context-aware attentional pooling to discover subtle variance that can characterizes the object. In \cite{ding2021ap}, Ding et al. propose a dual pathway hierarchy to integrate low-level information with high-level ones, which jointly utilizes spatial-wise and channel-wise attentions for discriminative regions extraction and feature fusion. 

Additionally, vision transformer (ViT) \cite{dosovitskiy2020image} and its variants that constituted with multi-head self-attention mechanism give another way to meet the challenge of fine-grained classification. By encoding image into patch sequence, the multi-head self-attention mechanism enables transformer layers in ViT to retain long-range dependency among these patches, and has achieved promising results. Many current fine-grained approaches are proposed based on this architecture \cite{hu2021rams,zhang2022free,rao2021counterfactual,he2022transfg,sun2022sim}. For example, RAMS-Trans \cite{hu2021rams} proposes dynamic patch proposal module to lead the region amplification of multi-scale learning. AF-Trans \cite{zhang2022free} accepts two-stage transformer structure that utilizes multi-scale information and filter them adaptively to capture region attention without box annotations. TransFG \cite{he2022transfg} proposes a part selection module to integrate all raw attention from transformer layers into an attention map that is able to guide the model to select discriminative patches. SIM-Trans \cite{sun2022sim} applies  polar coordinates to measure spatial relations among discriminative patches and the graph neural network is used to mine the object structure information, then the contrastive learning is used to enhancing feature robustness. 

More recently, Liu et al. \cite{liu2021swin} propose the Swin transformer (Swin-T), a hierarchical architecture using shifted window attention scheme. The Swin-T model is considered as a combination of CNN network and ViT, which has been proved effective to alleviate the limitations of ViT, such as small receptive field and large computing consumption. As a result, some Swin-T based methods have been proposed \cite{miao2021complemental,chou2022novel} for the FGVC task. In CAMF \cite{miao2021complemental}, Zhuang et al. enhance regional information for FGVC task by suppressing the most significant region. Chou et al. propose the PIM \cite{chou2022novel} to select useful features by finding out relatively unimportant features. However, the above methods pay little attention to the intrinsic challenge of FGVC that the similarity of subclasses can also be salient, which is the main reason of model's misclassification on FGVC task. Focusing on this challenge, our SFI-Net compares the most similar subcategories to locate and remove ambiguity regions for discriminative enhancement.

\begin{figure}[h]%
\centering
\includegraphics[width=1.0\textwidth]{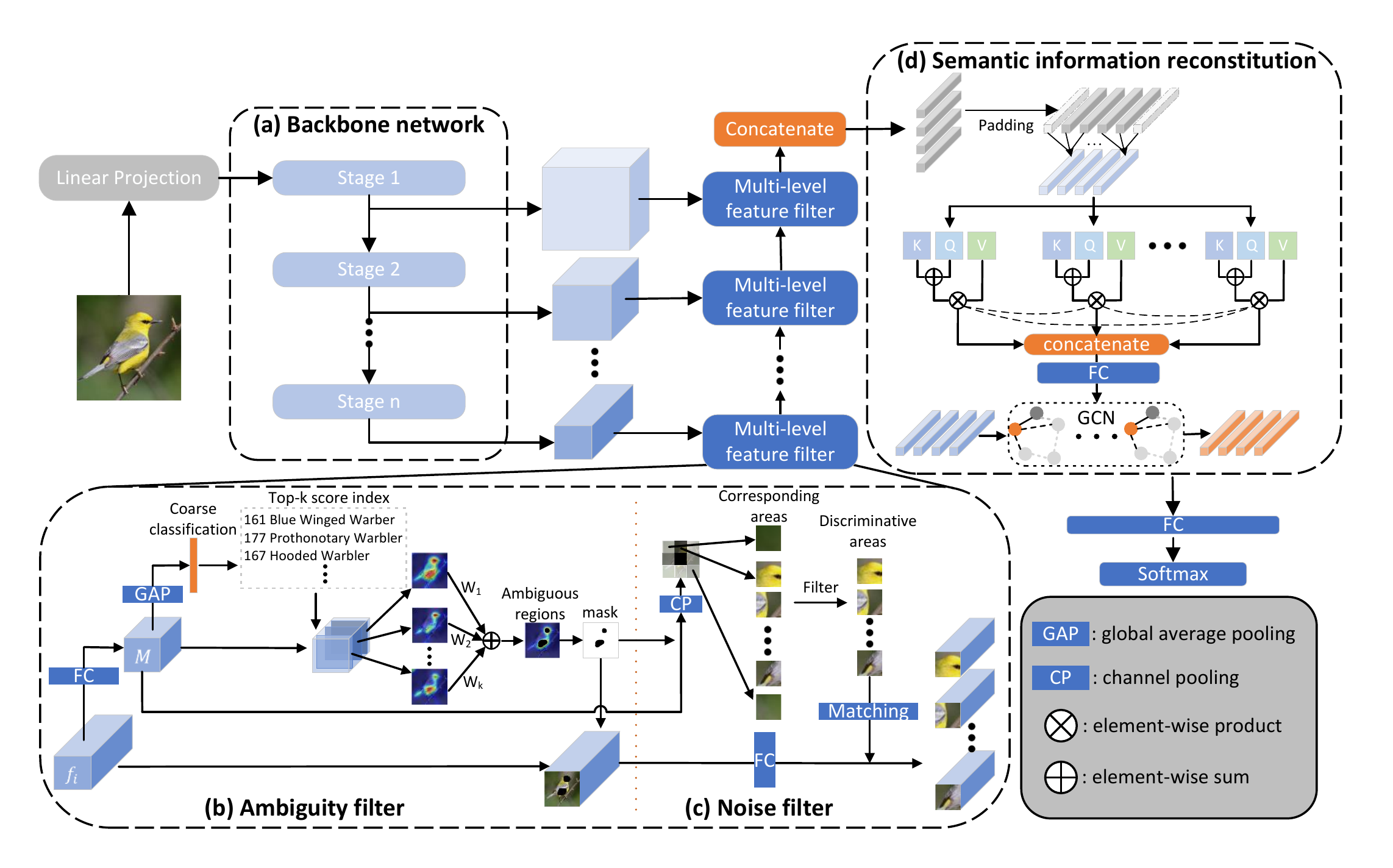}
\caption{The overall framefork of our SFI-net. The backbone network (a) uses Swin-T \cite{liu2021swin} to extract image feature. Ambiguity filter (b) and noise filter (c) are the two components of the multi-level feature filter (MFF) module. (d) is the semantic information reconstitution (SIR) module.}\label{fig2}
\end{figure}

\section{Method}\label{sec3}
Our proposed Structure Feature Integration Network 
(SFI-Net) attempts to optimize and then reconstruct features extracted from the backbone network for specializing in fine-grained visual classification task. The framework of our SFI-net is presented in Fig.~\ref{fig2}. We adopt Swin-T \cite{liu2021swin} as the backbone network (Fig.~\ref{fig2}(a)). For an input image, we first obtain its features of different stages of the backbone, then these features with different receptive fields are sent into our novel multi-level feature filter (MFF) module to remove unnecessary features. As shown in Fig.~\ref{fig2}(b) and (c), the MFF module consists of two sub-modules: ambiguity filter and noise filter. The semantic information reconstitution (SIR) module (Fig.~\ref{fig2}(d)) receives these retained discriminative features to reconstruct their contextual relationship for deciding the final classification.

\subsection{Multi-level Feature Filter}\label{subsec2}
The proposed multi-level feature filter module (MFF) consists of two sub-modules to screen out discriminative regions at pixel-level: 1) ambiguity filter (AF) which aims at finding out the most similar subcategories and eliminating unnecessary regions that share the striking resemblance; 2) noise filter (NF) which is adopted to discard noninformative features so that the negative influence of background noise can be alleviated to some extent.

\subsubsection{Ambiguity filter}\label{subsubsec2}

\begin{figure}[h]%
\centering
\includegraphics[width=1.0\textwidth]{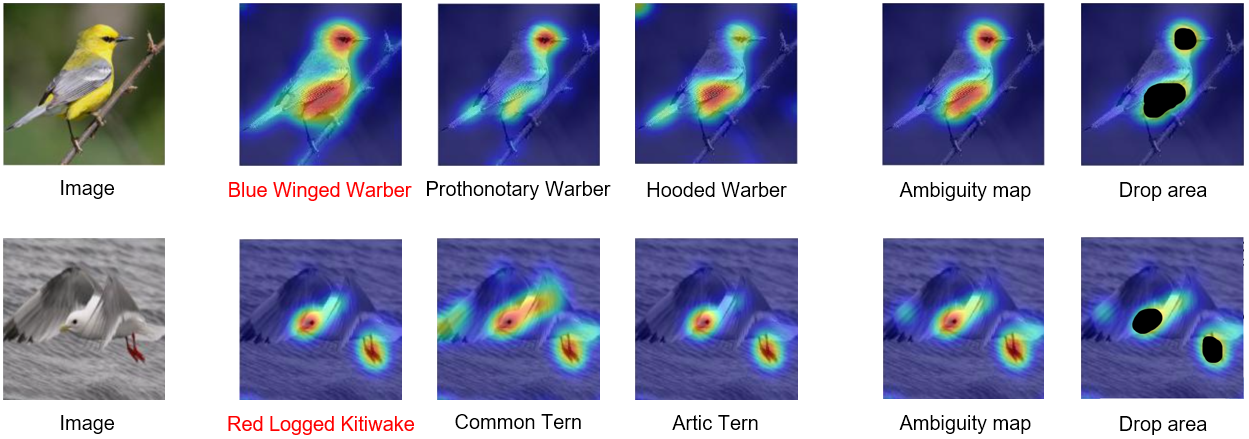}
\caption{Visual illustration of the process of our ambiguity filter module. The first column is input image, the second to fourth column are visualizations of the top-3 class feature maps, in which the label marked in red is the ground-truth. We select out the top-3 categories and calculate their similar region in the ambiguity map, i.e., the fifth column. The most ambiguous parts are dropped as shown in the sixth column, where regions marked in black denote the dropped areas.}\label{fig3}
\end{figure}

As shown in Fig.~\ref{fig2}(b), we propose a simple yet effective ambiguity filter (AF) module to resolve the ambiguity of top-k prediction. Our model takes an image $I$ as input. Firstly, $I$ is fed into the backbone network to generate feature map with different receptive fields. The output feature map of the $i^{th}$ stage is indicated as $f_i \in \mathbb{R}^{W\times H\times C}$, where $W$, $H$ and $C$ denote the width, height and channel dimension of feature map, respectively. Then, the feature map $f_i$ is sent into the module AF as input. Inspired by \cite{do2022fine}, we first obtain the class feature maps $M$ by applying the fully connected layer, where $M\in \mathbb{R}^{W\times H\times N}$ and $N$ denotes the number of target classes. The class feature maps $M$ is considered as the characterization of image $I$ on different class labels. Then, we utilize the global average pooling (GAP) to output coarse-grained prediction $p$ of image $I$ as follows:
\begin{equation}
p = GAP(M_N) = \frac{1}{H\times W}\sum_{i=1}^{H}\sum_{j=1}^{W}M_N(i,j)\label{eq1}
\end{equation}
where $p\in \mathbb{R}^{1\times N}$, $\left \{p_1,p_2,...,p_N\right\}$ denote $N$ prediction scores in $p$. Each score $p_i$ represents the relevance of $i$th category to the ground truth. We sort the prediction scores and find the top-k prediction feature maps $\left\{T_1,T_2,...,T_k\right\}$ along with their corresponding index in $p$ accordingly. $T\in \mathbb{R}^{W\times H\times k}$ is a subset of class feature maps $M$ and $k$ is a hyper-parameter used to control how many feature maps we choose. Then, we assign a set of weight coefficients $w = \left\{w_1,w_2,...,w_k\right\}$ with equal interval to represent the contribution importance of $T_i$. The interval and corresponding weight can be defined as:
\begin{align}
\delta = (\beta_h-\beta_l)/(k-1) \nonumber \\ w_i = \beta_h-(i-1)*\delta \label{eq2}
\end{align}
where $\delta$ is a interval between adjacent weight coefficients $w_i$ and $w_{i+1}$, $\beta_h$ and $\beta_l$ are the upper and lower limits of $w_i$. In order to filter out ambiguity regions, we do a weighted average on $T_k$, and the output ambiguity map $\mathcal{M}\in \mathbb{R}^{W\times H\times 1}$ is calculated as follows:
\begin{equation}
\mathcal{M} = \frac{1}{k}\sum_{i=1}^{k}w_iT_i\label{eq3}
\end{equation}
In order to filter out the ambiguity regions, we set a hyper-parameter $\gamma_1 $ as the ratio of ambiguity drop. The specific process is shown as follows:
\begin{equation}
\mathcal{M}_{(x,y)}'=\left\{\begin{array}{cc}
1, & \text { if } rank(\mathcal{M}_{(x, y)})<(1-\gamma_1)\times W\times H \\
0, & \text { otherwise }
\end{array}\right. \label{eq4}
\end{equation}
\begin{align}
M' = \mathcal{M}' \odot M \nonumber \\ f_i'=\mathcal{M}'\odot f_i\label{eq5}
\end{align}
where $\mathcal{M}'\in \mathbb{R}^{W\times H}$ denotes the filter mask for ambiguity drop; $\odot$ denotes Hadamard product; $\mathcal{M}_{(x,y)}$ denotes the score of $\mathcal{M}$ on position $(x,y)$; $M'$ is the class feature maps after ambiguity drop; $f_i'$ is the output of module AF. To maintain the stability of our model, we remove ambiguity regions with a fixed ratio $\gamma_1$, so $(1-\gamma_1)\times W\times H$ denotes the number of feature point we want to preserve. When the rank of $\mathcal{M}_{(x,y)}$ (i.e., $rank(\mathcal{M}_{(x, y)}$) is smaller than $(1-\gamma_1)\times W\times H$, the corresponding point is thought belonging to discriminative ones and will be preserved. Otherwise, the point will be set 0 to eliminate its effect.
In \cite{do2022fine}, Do et al. simply apply their dropping mechanism on original image, which results in a two-stage model. Different from that, our AF module processes features extracted from the backbone to output discriminative information, which can be trained end-to-end efficiently. Moreover, we utilize a set of weight coefficients to assign matching weights for feature maps with different scores. This step compensates for the unbalanced weights in \cite{do2022fine}. The visualization of each step in AF module is shown in Fig.~\ref{fig3}. 

From Fig.~\ref{fig3}, we can observe that the ambiguity region can be the most salient region (e.g. birds' head), which can intuitively hurt the performance of our model. However, we find that this phenomenon can stimulate our model to focus on other regional information, which results in a multiple attention extraction within the object extent. Experiments have proved the effectiveness of the AF module.

\subsubsection{Noise filter}\label{subsubsec3}
Although the AF module alleviates the degradation of model performance caused by ambiguity parts, there are still noninformative regions exist in image background. In order to reinforce the priority of discriminative regions, inspired by \cite{chou2022novel}, we introduce the noise filter (NF) module to screen out discriminative regions on pixel-level. The output of module AF $f_i'\in \mathbb{R}^{W\times H\times C}$ is applied as input of our NF module to eliminate background noise. We first obtain the spatial index map $A \in \mathbb{R}^{W\times H}$ by utilizing the channel average pooling (CP in Fig.~\ref{fig2}(c)) to compress the channel dimension of class feature maps $M'\in \mathbb{R}^{W\times H\times N}$. The spatial index map $A$ is used to represent reaction scores of pixels on spatial dimension. Each pixel on the spatial index map $A$ denotes a corresponding areas of image $I$ as shown in Fig.~\ref{fig2}(c). In \cite{chou2022novel}, Chou et al. use the maximal score to represent each channel. However, our experiment shows that maximal channel scores can lead to unstable training. In order to overcome this problem, we exploit the channel average score for a comprehensive characterization of each channel. Then, the spatial index map $A$ is flattened into a one-dimension vector $\mathcal{R}\in \mathbb{R}^{S}$, where $S = W\times H$. Meanwhile, the output of AF $f_i'$ is also flattened in spatial dimension from $(W\times H \times C)$ to $(S\times C)$ through fully connected layer. After this step, we compare each pixel score $\mathcal{R}_i, i=1,2,...,S$ in $\mathcal{R}$ with a specified threshold $\gamma_2$, where $\gamma_2$ is a hyper-parameter to control the length of output. A pixel will be discarded as background noise if its score rank is smaller than the rank of  $(1-\gamma_2)\times S$. As a result, we obtain the index sequence of selected features $\mathcal{R}'\in \mathbb{R}^{S'}$, where $S'$ denotes the length of $\mathcal{R}'$ and numerically equals to $(1-\gamma_2)\times S$. Finally, the discriminative descriptor $\mathcal{R}'$ is used to find corresponding regions on feature map $f_i'$ according to their spatial mapping relationships. $g\in \mathbb{R}^{S'\times C}$ denotes the preserved features, which is the output of module MFF.

During training, we use the cross-entropy loss to supervise MFF. In order to minimize computing expenses, we first obtain the average score of $g$, then we calculate the loss function $\mathcal{L}_{filt}$ as follows:
\begin{equation}
z_l = \frac{1}{S'}\sum_{i=1}^{S'}g_{(l,i)}  \label{eq6}
\end{equation}
\begin{equation}
\mathcal{L}_{filt} = -\sum_{l=1}^Lylog(pred(z_l)) \label{eq7}
\end{equation}
where $L$ is the number of stages of the backbone, $z_l$ is the average score of all features in $g$, $y$ is the one-hot label of ground truth and $pred(z_l)$ denotes the prediction vector of $z_l$.

Overall, our multi-level feature filter module is able to eliminate unnecessary features for discriminative information mining, which promotes the feature attention learning of our model.

\subsection{Semantic information reconstitution}\label{subsec3}
Although the output of MFF $g\in \mathbb{R}^{S'\times C}$ consists of discriminative features, the contextual connections among them are lost due to the two-level filter in MFF. These connections are of great importance for the overall cognition of objects. In order to make better use of these effective features, we introduce the semantic information reconstitution (SIR) module to gradually reconstruct the semantic relation of these preserved features. 

We first concatenate the output of MFF $g$ from different backbone stages, $G \in \mathbb{R}^{\mathcal{S}\times C}$ is the cocatenated features, where $\mathcal{S}$ denotes the number of all preserved features and $C$ is the channel dimension. After that, as shown in Fig.~\ref{fig2}(d), we adapt the semantic reassembly (SR) layer \cite{wang2021dynamic} to reconstruct the relation between spatially adjacent parts in $G$, where the padding operation is used to maintain the size of $G$. The specific formula is defined as follows:
\begin{equation}
B_i = \sum_{j=-1}^1W_j \otimes G_{i+j}\label{eq8}
\end{equation}
where the adjacent joint part $B_i\in\mathbb{R}^{C}$ is calculated through a method similar to one-dimensional convolution, in which we utilize element-wise multiplication $\otimes$ between the adjacent part $G_{i+j}$ and the corresponding learnable weight $W_j\in \mathbb{R}^C$. By applying this procedure, the neighboring content information is incorporated into the spatial isolated part $G_{i}$, and each neighboring part makes different contribution for the adjacent joint part $B_i$ due to variable semantic correlations.

The multi-headed self-attention (MSA) mechanism \cite{vaswani2017attention} is proved effective in establishing global connections of features, the attention weight between two pixels depict the strength of their correlation. Let's suppose the MSA has H heads, Q, K and V denote query matrix, key matrix and value matrix, respectively. The attention output in each head are calculated as follows:
\begin{equation}
J_h = softmax(\frac{QK^T}{\sqrt{C/H}})V\label{eq9}
\end{equation}
where $J_h\in \mathbb{R}^{\mathcal{S}\times (C/H)}$, $h = 1,2,...,H$ and $\mathcal{S}$ is the number of features in $G$. Matrices K, Q, and V are derived from $B$ by means of convolution. In \cite{vaswani2017attention}, the multi-headed self-attention (MSA) mechanism combines all attention outputs through directly concatenation, which causes 1) the separate computations in each head; and 2) the small receptive field limited by the size of attention head. To address these limitations, we adapt the talking-head attention (THA) \cite{shazeer2020talking} to connect each head. After obtaining the attention output $J_h$ in Eq.~\ref{eq9}, a learnable weight coefficient matrix $U$ is utilized to enable interaction between each head, which breaks the limitation of separate attention heads, and the specific formula is described as follows:
\begin{equation}
U = \left(\begin{array}{cccc}
\lambda_{11} & \lambda_{12} & \cdots & \lambda_{1 H} \\
\lambda_{21} & \lambda_{22} & \cdots & \lambda_{2 H} \\
\vdots & \vdots & \ddots & \vdots \\
\lambda_{H 1} & \lambda_{H 2} & \cdots & \lambda_{H H}
\end{array}\right)\label{eq10}
\end{equation}

\begin{equation}
\left(\begin{array}{c}
O_{1} \\
O_{2} \\
\vdots \\
O_{H}
\end{array}\right)=U\left(\begin{array}{c}
{J}_{1} \\
{J}_{2} \\
\vdots \\
{J}_{H}
\end{array}\right) 
\label{eq11}
\end{equation}
\begin{equation}
O = [O_{1},O_{2},...,O_{H}]\label{eq12}
\end{equation}
where the weight coefficient matrix $U\in \mathbb{R}^{H\times H}$ works as a linear projector for weights fusion, the output semantic features $O_h, h=1,2,...,H$ is incorporated with global information. After that, we concatenate these semantic features to obtain the semantic output $O\in \mathbb{R}^{\mathcal{S}\times C}$.

So far, we've established the semantic interaction of $\mathcal{S}$ discriminative patches, which is denoted as $O$. To characterize their spatial relationships, we design a graph convolutional network (GCN) \cite{kipf2016semi} to capture the object spatial structure. Firstly, a graph $G=(O,E)$ is conducted, which consists of two components: 1) the semantic features $O$ depict the importance of corresponding image regions, each feature $O_i\in \mathbb{R}^{C}$ in $O$ represents a node for subsequent integration; 2) the weighted edges $E$ are obtained through the backward propagation, which summarize the structure information among discriminative features within the object extent. Specifically, the adjacency matrix $Ad\in \mathbb{R}^{\mathcal{S}\times \mathcal{S}}$ denotes the strength of correlations among semantic features $O_i$. In order not to duplicate the relations established in SR and THA, we initialize $Ad$ to a fixed value and train it from scratch. During the backward propsagation, the edge weights vary based on the strength of correlations among semantic features $O_i$. And the graph convolution is applied to incorporate the structure information into $O$. The reconstituted features $f_{rec}$ are calculated as follows:
\begin{equation}
f_{rec}= \sigma(Ad\times O \times W)\label{eq13}
\end{equation}
where W is learnable parameter and $\sigma(\cdot)$ is activation function. Compared to integrating features in a single way, such as GNN in \cite{sun2022sim} and LSTM in \cite{behera2021context}, our SIR module gradually mine the semantic relations and spatial structure of discriminative features, which improves the semantic coherence of selected features through end-to-end training. We use the cross-entropy loss $\mathcal{L}_{cls}$ to supervise the final classification. The overall loss function $\mathcal{L}$ is represented as follows:
\begin{equation}
\mathcal{L}= \xi\mathcal{L}_{filt}+\mathcal{L}_{cls}\label{eq14}
\end{equation}
where $\xi$ is the balance coefficient.

\section{Experiments}\label{sec4}

In this section, we comprehensively evaluate our proposed SFI-Net on four widely used benchmark fine-grained visual classification datasets: CUB-200-2011 \cite{wah2011caltech},  Stanford Dogs \cite{khosla2011novel}, Stanford CArs \cite{krause20133d} and FGVC-aircraft \cite{maji2013fine}. We first introduce the datasets. Then, we present the implementation details of our experiments. Afterward, the comparative experiments, ablation study along with visualization study are conducted to demonstrate the effectiveness and accuracy of our proposed SFI-Net.

\begin{table}[h]
\begin{center}
\begin{minipage}{\textwidth}
\caption{Comparison with other advanced methods on CUB-200-2011 (CUB), Stanford-Dogs (Dogs), Stanford-Cars (Cars) and FGVC-aircraft (Airs). }  \label{tab1}%
\begin{tabular}{@{}llcccc@{}}
\toprule
Method & Year  & CUB Acc(\%) & Dogs Acc(\%) & Cars Acc(\%) & Airs Acc(\%)\\
\midrule
API-Net(AAAI) \cite{zhuang2020learning}    & 2020   & 90.0  & 88.1 & 89.4 & 93.9 \\
FDL(AAAI) \cite{liu2020filtration}    &  2020   & 88.6  & 85.0 & 94.2 & 93.4 \\
PMG(ECCV) \cite{du2020fine}   & 2020   & 89.6  & - & 95.1 & 93.4 \\
LIO(CVPR) \cite{zhou2020look}   &  2020  & 88.0 & - & 94.5 & 92.7 \\
DP-Net(AAAI) \cite{wang2021dynamic}  & 2021 & 89.3  & - & 94.8 & 93.9 \\
SEF(SPL) \cite{luo2020learning}    & 2020   &  87.3  & 88.8 & 94.0 & 92.1 \\
CAL(CVPR) \cite{rao2021counterfactual}   & 2021   & 90.6  & - & \underline{95.5} & \underline{94.2} \\
SA-MFN(APIN) \cite{chen2022fine}    & 2022   & 89.9  & 91.4 & 94.7 & 93.8 \\
CAP(AAAI) \cite{behera2021context}   & 2021   & 91.8  & - & \textbf{95.7} & \textbf{94.9} \\
SPS(ICCV) \cite{huang2021stochastic}   & 2021   & 88.7  & - & 94.9 & 92.7 \\

ELoPE(WACV) \cite{hanselmann2020elope}    & 2020   & 88.5  & - & 95.0 & 93.5 \\
ViT(ICLR) \cite{dosovitskiy2020image}   & 2020  & 90.6  & 91.7 & 93.7 & - \\
TransFG(AAAI) \cite{he2022transfg}   & 2022   & 91.7  & 92.3 & 94.8 & - \\
RAMS(ACMMM) \cite{hu2021rams}   & 2021   & 91.3  & 92.4 & - & - \\
DCAL(CVPR) \cite{zhu2022dual}   &  2022  & \underline{92.0}  & - & 95.3 & 93.3 \\
FFVT(BMVC) \cite{wang2021feature}   & 2021   & 91.6  & 91.5 & - & - \\

Swin-Trans(ICCV) \cite{liu2021swin}   & 2021   & 91.8  & 92.4 & 94.6 & 91.8 \\
CAMF(ISP) \cite{miao2021complemental}   & 2021   & 91.2  & \underline{92.8} & 95.3 & 93.3 \\
Ours    & -   & \textbf{92.6}  & \textbf{93.0} & 94.9 & 92.3 \\
\botrule
\end{tabular}
\end{minipage}
\end{center}
\end{table}

\subsection{Datasets}\label{subsec4}

Our experiments adopt four widely used FGVC datasets:
\begin{itemize}
\item CUB-200-2011 \cite{wah2011caltech} is published by California institute of technology, and is regarded as the most widely used dataset in the FGVC task. Including 11,788 images of 200 bird categories, where 5,994 images are for training and the rest 5,794 are used for testing.
\item Stanford Dogs \cite{khosla2011novel} has 20,580 images from 120 subclasses. Among them, 12,000 images are selected as training set and 8,580 images are selected as testing set.
\item Stanford Cars \cite{krause20133d} consists of 16,185 images spanning 196 classes, which is split into 8,144 training and 8,041 test images. This dataset is published by Stanford University.
\item FGVC-aircraft \cite{maji2013fine} contains 10,000 images of 100 different aircraft types. The training and testing set among them are divided by the ratio of 2:1.
\end{itemize} 

We use the top-1 accuracy (\%) as our evaluation metric and  comparison experiments are all conducted on these datasets.

\subsection{Implementation details}\label{subsec5}

In our experiments, Swin-T \cite{liu2021swin} is used as the backbone network, and the input image size is $384\times 384$. We follow recent works \cite{hu2021rams,sun2022sim,chou2022novel} to use standard FGVC pre-process steps: 1) resize the images into $510\times 510$; 2) randomly crop them into $384\times 384$ as the input size for training; 3) random horizontal flip and normalization are used. During testing phase, we resize the images into $510\times 510$ and crop them to $384\times 384$ using center crop. The balance coefficient $\xi$ in Eq.~\ref{eq14} is set as 3. We set the learning rate as 0.0005. The cosine decay is used and the weight decay is set to 0.0005. The stochastic gradient descent (SGD) optimizer is adopted with a momentum of 0.9. We set the batch size as 12, and the model is trained for 60 epochs on two NVIDIA Titan Xp GPUs.

\subsection{Comparison with state‑of‑the‑art methods}\label{subsec6}

In this section, we present the comparison results with previous state-of-the-art (SOTA) works over four FGVC datasets, which is shown in Table~\ref{tab1}. For fair comparison, all the works use no fine-grained annotations (e.g., object boxes and part locations). These SOTA methods include CNN based methods \cite{liu2020filtration,du2020fine,zhou2020look,wang2021dynamic,zhuang2020learning}, ViT based methods \cite{dosovitskiy2020image,he2022transfg,hu2021rams,zhu2022dual} and Swin-T based methods \cite{liu2021swin,miao2021complemental}. Among them, the classification accuracy of \cite{liu2021swin} is obtained through pre-trained models provided by its authors, and other accuracy results are obtained by relative references. 

From Table~\ref{tab1}, we can observe that there is no work can reach the best performance on all datasets over the past three years. Our proposed SFI-Net surpasses most of these SOTA methods with a convincing performance gain. To be more specific, Our work achieves 0.8\% and 1.6\% performance improvement on CUB-200-2011 \cite{wah2011caltech} and Stanford Dogs \cite{khosla2011novel} compared with the optimal CNN based method (i.e. \cite{behera2021context} and \cite{chen2022fine}). CNN based methods generally exploit attention mechanism and multi-scale learning to find discriminative information and construct their relations. For instance, jigsaw operation is introduced in PMG \cite{du2020fine} to mine mutli-granularity information, which enhances model's recognizing ability from different scales. Whereas, the performance of CNN based methods is inherent limited due to weak ability adapting to big pre-training data. As a result, many ViT based methods with outstanding performance are proposed, which demand huge pre-training dataset. Compared with them, our SFI-Net also shows advanced performance, attaining 0.6\% performance improvement compared with DCAL \cite{zhu2022dual} on CUB-200-2011 and 0.6\% performance gain on Stanford Dogs compared with RAMS-Trans \cite{hu2021rams}. Note that DCAL and RAMS-Trans are optimal ViT based methods on CUB-200-2011 and Stanford Dogs, respectively. 

More recently, Swin-Transformer \cite{liu2021swin} construct a hierarchical representation of patches by applying the shifted window scheme, which enables us to extract features from different scale while reducing the computational consumption. Thus, the Swin-T based methods have the advantage on FGVC for bigger receptive field than ViT based methods. As shown in Table~\ref{tab1}, our SFi-Net outperforms the CAMF \cite{miao2021complemental} by a margin of 1.4\% and 0.2\% on CUB-200-2011 and Stanford Dogs. We can also observe in Table~\ref{tab1} that our model achieves 0.8\%, 0.6\%, 0.3\% and 0.5\% accuracy improvement on CUB-200-2011, Stanford Dogs, Stanford Cars \cite{krause20133d}, and FGVC-aircraft \cite{maji2013fine} than the baseline network Swin-T, respectively.

\subsection{Ablation study}\label{subsec7}
We conduct ablation experiments on CUB-200-2011 in this section. We propose multi-level 
feature filter (MFF) module and semantic information reconstitution (SIR) module in our SFI-Net. Each module has several corresponding components and hyper-parameters. The ablation results are shown in Table~\ref{tab2}-\ref{tab6}, which can be divided into two parts: 1) component analysis and 2) hyper-parameter experiments.

\subsubsection{Component analysis}\label{subsubsec4}

\begin{table}[h]
\begin{center}
\begin{minipage}{180pt}
\caption{Ablation Experiments on CUB-200-2011 dataset}\label{tab2}%
\begin{tabular}{@{}ll@{}}
\toprule
Method & ACC(\%) \\
\midrule
Baseline    & 91.8     \\
Baseline + MFF\_without\_NF    & 92.0(+0.2)     \\
Baseline + MFF    & 92.3(+0.5)    \\
Baseline + MFF + SIR    & \textbf{92.6}(+0.8)     \\
\botrule
\end{tabular}

\end{minipage}
\end{center}
\end{table}

The contribution of each component in our SFI-Net is shown in Table~\ref{tab2}, we conduct different settings on CUB-200-2011 by utilizing Swin-L-384 with a classifier head as the baseline. Firstly, we evaluate the contribution of ambiguity filter (AF) by directly introducing AF (i.e. MFF without NF) to discard ambiguous regions, the top-1 recognition accuracy on CUB-200-2011 improves 0.2\%. Then, the MFF consisting of AF and NF is plugged into all the stages of the baseline as an entity, which brings an improvement of 0.5\% classification accuracy. Finally, we conduct the semantic information reconstitution (SIR) based on the MFF module, which improves the performance of our model with another 0.3\% accuracy gain.

\subsubsection{Hyper-parameter experiments}\label{subsubsec4}
In this section, we analyse four groups of hyper-parameter in our proposed SFI-Net.
\begin{itemize}
\item The hyper-parameter $k$ in Eq.~\ref{eq3} controls the number of top-k ambiguity feature maps we select in AF module. As shown in Table~\ref{tab3}, our model achieves the best classification accuracy when we set $k$ to be 4. If $k$ is set too small, these feature maps may not represent the ambiguity regions. While if $k$ is set too big, dissimilar categories can be involved.
\item The $\beta_h$ and $\beta_l$ are used to determine the upper and lower limits of the top-k feature maps' weight. We can observe from Table~\ref{tab4} that when $\beta_h = 1.1$ and $\beta_l = 0.95$, our model reaches the best accuracy. Although the classification accuracy does not change regularly as $\beta_h$ and $\beta_l$ fluctuate, a moderate interval between upper and lower limit can improve the model's performance. The reason may be that a large difference between $\beta_h$ and $\beta_l$
can not match the importance of each top-k feature map.
\item The $\gamma_1$ and $\gamma_2$ represent the drop rate of AF and NF. It is obvious in Table~\ref{tab5} that our model achieves the best performance with $\gamma_1=0.1$ and $\gamma_2=0.2$, respectively. If setting them bigger, the accuracy degrades because the number of features left is too small. On the opposite, setting them smaller makes too many features left, which can cause the remain of unnecessary noise. Note that in AF we simply set ambiguity regions to 0, and the corresponding removal operation is included in NF. So $\gamma_1$ can not be set bigger than $\gamma_2$.
\item Many previous works \cite{hu2021rams,sun2022sim,bera2022sr} have proved that using too many graph convolutional network (GCN) layers to pass message can cause over-smoothing, which harms the spatial structure construction within object scope. Our experiment in Table~ \ref{tab6} also verify this conclusion. As the depth of GCN increases, the performance of our model degrades due to losing the focus of local representation.
\end{itemize}

\begin{table}[h]
\begin{center}
\begin{minipage}{160pt}
\caption{Hyper-parameter analysis of K on CUB-200-2011}\label{tab3}%
\begin{tabular}{@{}lcccc@{}}
\toprule
K & 3  & 4 & 5 & 6  \\
\midrule
ACC(\%)    &  92.49  & \textbf{92.64}  & 92.46 & 92.53 \\

\botrule
\end{tabular}
\end{minipage}
\end{center}
\end{table}

\begin{table}[h]
\begin{center}
\begin{minipage}{200pt}
\caption{Hyper-parameter analysis of $\beta_1$ and $\beta_2$ on CUB-200-2011}\label{tab4}%
\begin{tabular}{@{}lcccc@{}}
\toprule
CUB & $\beta_h$=1.05  & $\beta_h$=1.1 & $\beta_h$=1.15 & $\beta_h$=1.2\\
\midrule
$\beta_l$=0.95    & 92.42\%   & \textbf{92.64\%}  & 92.30\% & 92.34\% \\
$\beta_l$=0.9    & 92.51\%   & 92.35\%  & 92.13\% & 92.21\%\\
$\beta_l$=0.85    & 92.41\%   & 92.32\%  & 92.57\% & 92.31\% \\
$\beta_l$=0.8    & 92.05\%   & 92.36\%  & 92.42\% & 92.38\%\\
\botrule
\end{tabular}

\end{minipage}
\end{center}
\end{table}

\begin{table}[h]
\begin{center}
\begin{minipage}{200pt}
\caption{Hyper-parameter analysis of $\gamma_1$ and $\gamma_2$ on CUB-200-2011}\label{tab5}%
\begin{tabular}{@{}lcccc@{}}
\toprule
CUB & $\gamma_1$=0.05  & $\gamma_1$=0.1 & $\gamma_1$=0.15 & $\gamma_1$=0.2\\
\midrule
$\gamma_2$=0.15    & 92.42   & 92.47  & 92.20 & - \\
$\gamma_2$=0.2    & 92.51   & \textbf{92.64}  & 92.13 & 91.68\\
$\gamma_2$=0.25    & 92.41   & 92.32  & 91.97 & 91.31\\
$\gamma_2$=0.3    & 92.05   & 91.76  & 91.32 & 90.98\\
\botrule
\end{tabular}

\end{minipage}
\end{center}
\end{table}

\begin{table}[h]
\begin{center}
\begin{minipage}{200pt}
\caption{Hyper-parameter analysis of GCN Depth on CUB-200-2011}\label{tab6}%
\begin{tabular}{@{}lccccc@{}}
\toprule
Depth & 1  & 2 & 3 & 4 & 5 \\
\midrule
ACC(\%)    & \textbf{92.64}   & 92.47  & 92.51 & 92.23 & 91.96\\

\botrule
\end{tabular}

\end{minipage}
\end{center}
\end{table}

\begin{figure}[h]%
\centering
\includegraphics[width=0.95\textwidth]{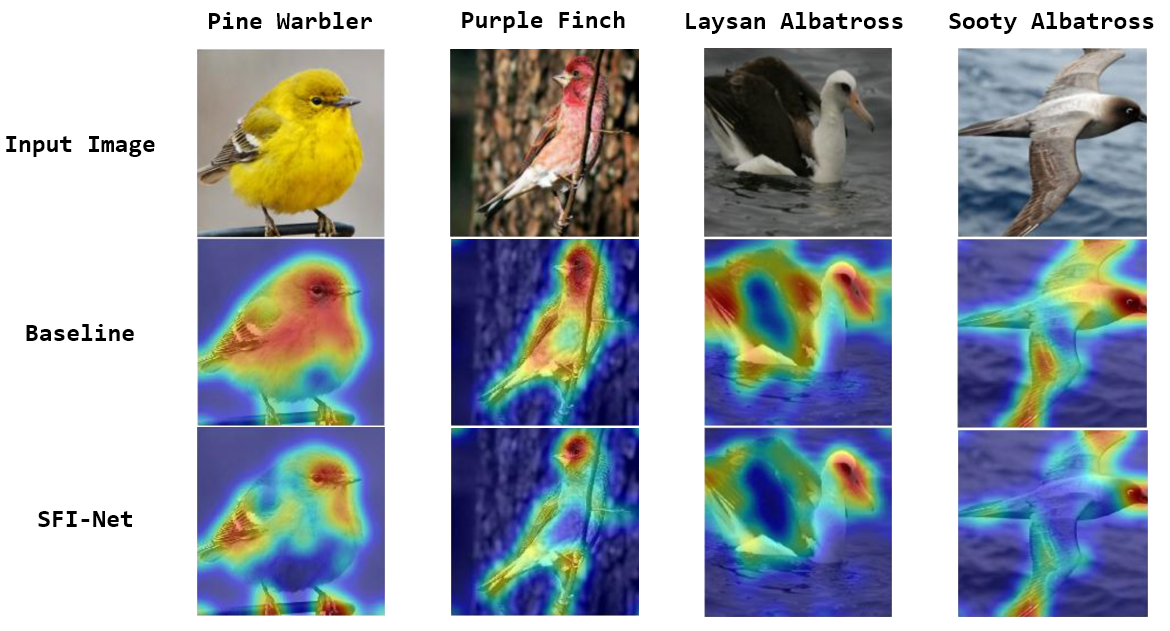}
\caption{Visualization of the ambiguity filtration results of our SFI-Net compared with the baseline. The first row is the original image and the second row is the visualization of feature maps obtained from the baseline. The last row is the visualization results of our SFI-Net. It can be seen that our SFI-Net can effectively remove ambiguous regions within the object extent, and the most discriminative regions are preserved compared with the baseline model \cite{liu2021swin}.}\label{fig4}
\end{figure}

\begin{figure}[h]%
\centering
\includegraphics[width=0.95\textwidth]{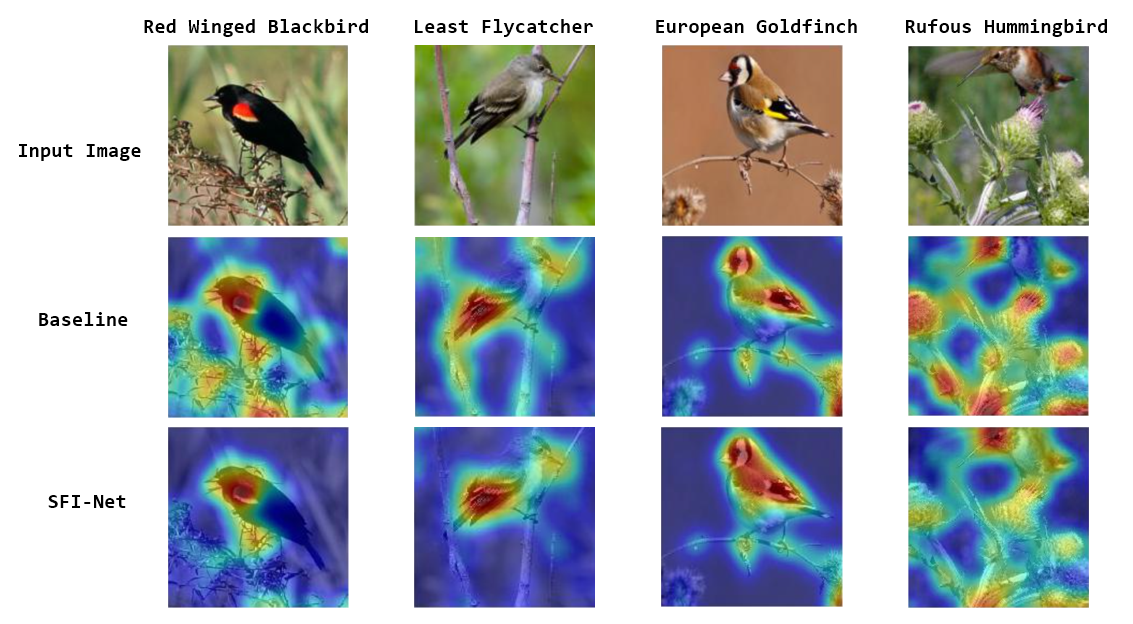}
\caption{Comparison of the denoising ability of our SFI-Net and the baseline. It is obvious that our SFI-Net surpasses the baseline on noise elimination. These four images are chosen from CUB-200-2011 and their corresponding heat maps are obtained through gradient-weighted class activation mapping \cite{selvaraju2017grad} technology.}\label{fig5}
\end{figure}

\subsection{Visualization Analysis}\label{subsec8}
The essential goal of our SFI-Net is to filter out 1) ambiguity regions that cause the confusion of our model leading to misdistinguishment and 2) background noise, which harms the understanding of object structure. Thus, we conduct respective visualization experiment to investigate the performance of our proposed SFI-Net.

As shown in Fig.~\ref{fig4}, we exploit the heat maps to analyse our model's ambiguity removal ability. The first row are original images from CUB-200-2011, the second row are heat maps obtained by the baseline and the last row are heat maps obtained by our SFI-Net, where the red highlighted areas represent the focus areas and the blue regions belong to weak focus areas. We randomly choose four bird images. It can be clearly seen that the original baseline without SFI-Net activates a large range of features within the object region, while many of these features are unimportant or ambiguous due to belonging to resemble parts. Our SFI-Net alleviates this problem effectively as shown in the third row of Fig.~\ref{fig4}. Compared with the baseline, ambiguity regions are visibly eliminated by our SFI-Net, which enhances the priority of discriminative regions (e.g. the head and the lower end of wings with distinctive feathers). It is also evident that our SFI-Net preserves the integrity of discriminative areas to the maximum, which is greatly helpful for deciding the final classification.

 In Fig.~\ref{fig5}, we demonstrate the effectiveness of our model's denoising ability. We randomly choose four bird images with strong background noise (e.g. twigs and flowers). It is clear in Fig.~\ref{fig5} that our SFI-Net can effectively remove or alleviate the impact of background noise and keeps the attention of our model concentrated to focus on the  discriminative regions within object extent compared with the baseline, which indicates our SFI-Net can learn and distinguish the difference between object's structure information and background noise.

\section{Conclusion}\label{sec5}
In this paper, we propose a novel semantic feature intergration network (SFI-Net) to preserve the most 
discriminative features for semantic information learning. The multi-level filter (MFF) module is introduced to eliminate unnecessary regions that have negative effect on the final classification, which increases the priority of discriminative features on pixel level. And the semantic information reconstitution (SIR) module is proposed to construct contextual relations among these scattered features, which makes better use of these preserved information. These two module promote each other and work as an entirety. By proposing the SFI-Net, we attempt to directly integrate the preserved feature from different stages and have obtained promising results. The superior performance of our SFI-Net is verified through extensive experiments on various FGVC datasets.

In the future, we will carry on the research of ambiguity regions elimination for better performance in FGVC task. Also, the contrastive learning will be studied to catch subtle changes among subcategories for a more robust training.

\section{Declarations}\label{sec6}
\textbf{Conflict of Interests }  
The authors declare no conflict of interest.

\bibliography{sn-bibliography}


\begin{thebibliography}{45}
\ifx \bisbn   \undefined \def \bisbn  #1{ISBN #1}\fi
\ifx \binits  \undefined \def \binits#1{#1}\fi
\ifx \bauthor  \undefined \def \bauthor#1{#1}\fi
\ifx \batitle  \undefined \def \batitle#1{#1}\fi
\ifx \bjtitle  \undefined \def \bjtitle#1{#1}\fi
\ifx \bvolume  \undefined \def \bvolume#1{\textbf{#1}}\fi
\ifx \byear  \undefined \def \byear#1{#1}\fi
\ifx \bissue  \undefined \def \bissue#1{#1}\fi
\ifx \bfpage  \undefined \def \bfpage#1{#1}\fi
\ifx \blpage  \undefined \def \blpage #1{#1}\fi
\ifx \burl  \undefined \def \burl#1{\textsf{#1}}\fi
\ifx \doiurl  \undefined \def \doiurl#1{\url{https://doi.org/#1}}\fi
\ifx \betal  \undefined \def \betal{\textit{et al.}}\fi
\ifx \binstitute  \undefined \def \binstitute#1{#1}\fi
\ifx \binstitutionaled  \undefined \def \binstitutionaled#1{#1}\fi
\ifx \bctitle  \undefined \def \bctitle#1{#1}\fi
\ifx \beditor  \undefined \def \beditor#1{#1}\fi
\ifx \bpublisher  \undefined \def \bpublisher#1{#1}\fi
\ifx \bbtitle  \undefined \def \bbtitle#1{#1}\fi
\ifx \bedition  \undefined \def \bedition#1{#1}\fi
\ifx \bseriesno  \undefined \def \bseriesno#1{#1}\fi
\ifx \blocation  \undefined \def \blocation#1{#1}\fi
\ifx \bsertitle  \undefined \def \bsertitle#1{#1}\fi
\ifx \bsnm \undefined \def \bsnm#1{#1}\fi
\ifx \bsuffix \undefined \def \bsuffix#1{#1}\fi
\ifx \bparticle \undefined \def \bparticle#1{#1}\fi
\ifx \barticle \undefined \def \barticle#1{#1}\fi
\bibcommenthead
\ifx \bconfdate \undefined \def \bconfdate #1{#1}\fi
\ifx \botherref \undefined \def \botherref #1{#1}\fi
\ifx \url \undefined \def \url#1{\textsf{#1}}\fi
\ifx \bchapter \undefined \def \bchapter#1{#1}\fi
\ifx \bbook \undefined \def \bbook#1{#1}\fi
\ifx \bcomment \undefined \def \bcomment#1{#1}\fi
\ifx \oauthor \undefined \def \oauthor#1{#1}\fi
\ifx \citeauthoryear \undefined \def \citeauthoryear#1{#1}\fi
\ifx \endbibitem  \undefined \def \endbibitem {}\fi
\ifx \bconflocation  \undefined \def \bconflocation#1{#1}\fi
\ifx \arxivurl  \undefined \def \arxivurl#1{\textsf{#1}}\fi
\csname PreBibitemsHook\endcsname

\bibitem{liu2021swin}
\begin{bchapter}
\bauthor{\bsnm{Liu}, \binits{Z.}},
\bauthor{\bsnm{Lin}, \binits{Y.}},
\bauthor{\bsnm{Cao}, \binits{Y.}},
\bauthor{\bsnm{Hu}, \binits{H.}},
\bauthor{\bsnm{Wei}, \binits{Y.}},
\bauthor{\bsnm{Zhang}, \binits{Z.}},
\bauthor{\bsnm{Lin}, \binits{S.}},
\bauthor{\bsnm{Guo}, \binits{B.}}:
\bctitle{Swin transformer: Hierarchical vision transformer using shifted
  windows}.
In: \bbtitle{Proceedings of the IEEE/CVF International Conference on Computer
  Vision},
pp. \bfpage{10012}--\blpage{10022}
(\byear{2021})
\end{bchapter}
\endbibitem

\bibitem{selvaraju2017grad}
\begin{bchapter}
\bauthor{\bsnm{Selvaraju}, \binits{R.R.}},
\bauthor{\bsnm{Cogswell}, \binits{M.}},
\bauthor{\bsnm{Das}, \binits{A.}},
\bauthor{\bsnm{Vedantam}, \binits{R.}},
\bauthor{\bsnm{Parikh}, \binits{D.}},
\bauthor{\bsnm{Batra}, \binits{D.}}:
\bctitle{Grad-cam: Visual explanations from deep networks via gradient-based
  localization}.
In: \bbtitle{Proceedings of the IEEE International Conference on Computer
  Vision},
pp. \bfpage{618}--\blpage{626}
(\byear{2017})
\end{bchapter}
\endbibitem

\bibitem{zhang2014part}
\begin{bchapter}
\bauthor{\bsnm{Zhang}, \binits{N.}},
\bauthor{\bsnm{Donahue}, \binits{J.}},
\bauthor{\bsnm{Girshick}, \binits{R.}},
\bauthor{\bsnm{Darrell}, \binits{T.}}:
\bctitle{Part-based r-cnns for fine-grained category detection}.
In: \bbtitle{European Conference on Computer Vision},
pp. \bfpage{834}--\blpage{849}
(\byear{2014}).
\bcomment{Springer}
\end{bchapter}
\endbibitem

\bibitem{wei2016mask}
\begin{botherref}
\oauthor{\bsnm{Wei}, \binits{X.-S.}},
\oauthor{\bsnm{Xie}, \binits{C.-W.}},
\oauthor{\bsnm{Wu}, \binits{J.}}:
Mask-cnn: Localizing parts and selecting descriptors for fine-grained image
  recognition.
arXiv preprint arXiv:1605.06878
(2016)
\end{botherref}
\endbibitem

\bibitem{branson2014bird}
\begin{botherref}
\oauthor{\bsnm{Branson}, \binits{S.}},
\oauthor{\bsnm{Van~Horn}, \binits{G.}},
\oauthor{\bsnm{Belongie}, \binits{S.}},
\oauthor{\bsnm{Perona}, \binits{P.}}:
Bird species categorization using pose normalized deep convolutional nets.
arXiv preprint arXiv:1406.2952
(2014)
\end{botherref}
\endbibitem

\bibitem{berg2013poof}
\begin{bchapter}
\bauthor{\bsnm{Berg}, \binits{T.}},
\bauthor{\bsnm{Belhumeur}, \binits{P.N.}}:
\bctitle{Poof: Part-based one-vs.-one features for fine-grained categorization,
  face verification, and attribute estimation}.
In: \bbtitle{Proceedings of the IEEE Conference on Computer Vision and Pattern
  Recognition},
pp. \bfpage{955}--\blpage{962}
(\byear{2013})
\end{bchapter}
\endbibitem

\bibitem{huang2016part}
\begin{bchapter}
\bauthor{\bsnm{Huang}, \binits{S.}},
\bauthor{\bsnm{Xu}, \binits{Z.}},
\bauthor{\bsnm{Tao}, \binits{D.}},
\bauthor{\bsnm{Zhang}, \binits{Y.}}:
\bctitle{Part-stacked cnn for fine-grained visual categorization}.
In: \bbtitle{Proceedings of the IEEE Conference on Computer Vision and Pattern
  Recognition},
pp. \bfpage{1173}--\blpage{1182}
(\byear{2016})
\end{bchapter}
\endbibitem

\bibitem{bib13}
\begin{botherref}
\oauthor{\bsnm{Do}, \binits{T.}},
\oauthor{\bsnm{Tran}, \binits{H.}},
\oauthor{\bsnm{Tjiputra}, \binits{E.}},
\oauthor{\bsnm{Tran}, \binits{Q.D.}},
\oauthor{\bsnm{Nguyen}, \binits{A.}}:
Fine-grained visual classification using self assessment classifier.
arXiv:2205.10529
(2022)
\end{botherref}
\endbibitem

\bibitem{lin2015bilinear}
\begin{bchapter}
\bauthor{\bsnm{Lin}, \binits{T.-Y.}},
\bauthor{\bsnm{RoyChowdhury}, \binits{A.}},
\bauthor{\bsnm{Maji}, \binits{S.}}:
\bctitle{Bilinear cnn models for fine-grained visual recognition}.
In: \bbtitle{Proceedings of the IEEE International Conference on Computer
  Vision},
pp. \bfpage{1449}--\blpage{1457}
(\byear{2015})
\end{bchapter}
\endbibitem

\bibitem{zhuang2020learning}
\begin{bchapter}
\bauthor{\bsnm{Zhuang}, \binits{P.}},
\bauthor{\bsnm{Wang}, \binits{Y.}},
\bauthor{\bsnm{Qiao}, \binits{Y.}}:
\bctitle{Learning attentive pairwise interaction for fine-grained
  classification}.
In: \bbtitle{Proceedings of the AAAI Conference on Artificial Intelligence},
vol. \bseriesno{34},
pp. \bfpage{13130}--\blpage{13137}
(\byear{2020})
\end{bchapter}
\endbibitem

\bibitem{li2020attribute}
\begin{bchapter}
\bauthor{\bsnm{Li}, \binits{H.}},
\bauthor{\bsnm{Zhang}, \binits{X.}},
\bauthor{\bsnm{Tian}, \binits{Q.}},
\bauthor{\bsnm{Xiong}, \binits{H.}}:
\bctitle{Attribute mix: Semantic data augmentation for fine grained
  recognition}.
In: \bbtitle{2020 IEEE International Conference on Visual Communications and
  Image Processing (VCIP)},
pp. \bfpage{243}--\blpage{246}
(\byear{2020}).
\bcomment{IEEE}
\end{bchapter}
\endbibitem

\bibitem{du2020fine}
\begin{bchapter}
\bauthor{\bsnm{Du}, \binits{R.}},
\bauthor{\bsnm{Chang}, \binits{D.}},
\bauthor{\bsnm{Bhunia}, \binits{A.K.}},
\bauthor{\bsnm{Xie}, \binits{J.}},
\bauthor{\bsnm{Ma}, \binits{Z.}},
\bauthor{\bsnm{Song}, \binits{Y.-Z.}},
\bauthor{\bsnm{Guo}, \binits{J.}}:
\bctitle{Fine-grained visual classification via progressive multi-granularity
  training of jigsaw patches}.
In: \bbtitle{European Conference on Computer Vision},
pp. \bfpage{153}--\blpage{168}
(\byear{2020}).
\bcomment{Springer}
\end{bchapter}
\endbibitem

\bibitem{liu2021subtler}
\begin{barticle}
\bauthor{\bsnm{Liu}, \binits{C.}},
\bauthor{\bsnm{Huang}, \binits{L.}},
\bauthor{\bsnm{Wei}, \binits{Z.}},
\bauthor{\bsnm{Zhang}, \binits{W.}}:
\batitle{Subtler mixed attention network on fine-grained image classification}.
\bjtitle{Applied Intelligence}
\bvolume{51}(\bissue{11}),
\bfpage{7903}--\blpage{7916}
(\byear{2021})
\end{barticle}
\endbibitem

\bibitem{behera2021context}
\begin{bchapter}
\bauthor{\bsnm{Behera}, \binits{A.}},
\bauthor{\bsnm{Wharton}, \binits{Z.}},
\bauthor{\bsnm{Hewage}, \binits{P.R.}},
\bauthor{\bsnm{Bera}, \binits{A.}}:
\bctitle{Context-aware attentional pooling (cap) for fine-grained visual
  classification}.
In: \bbtitle{Proceedings of the AAAI Conference on Artificial Intelligence},
vol. \bseriesno{35},
pp. \bfpage{929}--\blpage{937}
(\byear{2021})
\end{bchapter}
\endbibitem

\bibitem{fu2017look}
\begin{bchapter}
\bauthor{\bsnm{Fu}, \binits{J.}},
\bauthor{\bsnm{Zheng}, \binits{H.}},
\bauthor{\bsnm{Mei}, \binits{T.}}:
\bctitle{Look closer to see better: Recurrent attention convolutional neural
  network for fine-grained image recognition}.
In: \bbtitle{Proceedings of the IEEE Conference on Computer Vision and Pattern
  Recognition},
pp. \bfpage{4438}--\blpage{4446}
(\byear{2017})
\end{bchapter}
\endbibitem

\bibitem{ding2019selective}
\begin{bchapter}
\bauthor{\bsnm{Ding}, \binits{Y.}},
\bauthor{\bsnm{Zhou}, \binits{Y.}},
\bauthor{\bsnm{Zhu}, \binits{Y.}},
\bauthor{\bsnm{Ye}, \binits{Q.}},
\bauthor{\bsnm{Jiao}, \binits{J.}}:
\bctitle{Selective sparse sampling for fine-grained image recognition}.
In: \bbtitle{Proceedings of the IEEE/CVF International Conference on Computer
  Vision},
pp. \bfpage{6599}--\blpage{6608}
(\byear{2019})
\end{bchapter}
\endbibitem

\bibitem{chen2022fine}
\begin{botherref}
\oauthor{\bsnm{Chen}, \binits{H.}},
\oauthor{\bsnm{Cheng}, \binits{L.}},
\oauthor{\bsnm{Huang}, \binits{G.}},
\oauthor{\bsnm{Zhang}, \binits{G.}},
\oauthor{\bsnm{Lan}, \binits{J.}},
\oauthor{\bsnm{Yu}, \binits{Z.}},
\oauthor{\bsnm{Pun}, \binits{C.-M.}},
\oauthor{\bsnm{Ling}, \binits{W.-K.}}:
Fine-grained visual classification with multi-scale features based on
  self-supervised attention filtering mechanism.
Applied Intelligence,
1--17
(2022)
\end{botherref}
\endbibitem

\bibitem{liu2020filtration}
\begin{bchapter}
\bauthor{\bsnm{Liu}, \binits{C.}},
\bauthor{\bsnm{Xie}, \binits{H.}},
\bauthor{\bsnm{Zha}, \binits{Z.-J.}},
\bauthor{\bsnm{Ma}, \binits{L.}},
\bauthor{\bsnm{Yu}, \binits{L.}},
\bauthor{\bsnm{Zhang}, \binits{Y.}}:
\bctitle{Filtration and distillation: Enhancing region attention for
  fine-grained visual categorization}.
In: \bbtitle{Proceedings of the AAAI Conference on Artificial Intelligence},
vol. \bseriesno{34},
pp. \bfpage{11555}--\blpage{11562}
(\byear{2020})
\end{bchapter}
\endbibitem

\bibitem{hanselmann2020elope}
\begin{bchapter}
\bauthor{\bsnm{Hanselmann}, \binits{H.}},
\bauthor{\bsnm{Ney}, \binits{H.}}:
\bctitle{Elope: Fine-grained visual classification with efficient localization,
  pooling and embedding}.
In: \bbtitle{Proceedings of the IEEE/CVF Winter Conference on Applications of
  Computer Vision},
pp. \bfpage{1247}--\blpage{1256}
(\byear{2020})
\end{bchapter}
\endbibitem

\bibitem{wah2011caltech}
\begin{botherref}
\oauthor{\bsnm{Wah}, \binits{C.}},
\oauthor{\bsnm{Branson}, \binits{S.}},
\oauthor{\bsnm{Welinder}, \binits{P.}},
\oauthor{\bsnm{Perona}, \binits{P.}},
\oauthor{\bsnm{Belongie}, \binits{S.}}:
The caltech-ucsd birds-200-2011 dataset
(2011)
\end{botherref}
\endbibitem

\bibitem{khosla2011novel}
\begin{bchapter}
\bauthor{\bsnm{Khosla}, \binits{A.}},
\bauthor{\bsnm{Jayadevaprakash}, \binits{N.}},
\bauthor{\bsnm{Yao}, \binits{B.}},
\bauthor{\bsnm{Li}, \binits{F.-F.}}:
\bctitle{Novel dataset for fine-grained image categorization: Stanford dogs}.
In: \bbtitle{Proc. CVPR Workshop on Fine-grained Visual Categorization (FGVC)},
vol. \bseriesno{2}
(\byear{2011}).
\bcomment{Citeseer}
\end{bchapter}
\endbibitem

\bibitem{krause20133d}
\begin{bchapter}
\bauthor{\bsnm{Krause}, \binits{J.}},
\bauthor{\bsnm{Stark}, \binits{M.}},
\bauthor{\bsnm{Deng}, \binits{J.}},
\bauthor{\bsnm{Fei-Fei}, \binits{L.}}:
\bctitle{3d object representations for fine-grained categorization}.
In: \bbtitle{Proceedings of the IEEE International Conference on Computer
  Vision Workshops},
pp. \bfpage{554}--\blpage{561}
(\byear{2013})
\end{bchapter}
\endbibitem

\bibitem{maji2013fine}
\begin{botherref}
\oauthor{\bsnm{Maji}, \binits{S.}},
\oauthor{\bsnm{Rahtu}, \binits{E.}},
\oauthor{\bsnm{Kannala}, \binits{J.}},
\oauthor{\bsnm{Blaschko}, \binits{M.}},
\oauthor{\bsnm{Vedaldi}, \binits{A.}}:
Fine-grained visual classification of aircraft.
arXiv preprint arXiv:1306.5151
(2013)
\end{botherref}
\endbibitem

\bibitem{huang2021snapmix}
\begin{bchapter}
\bauthor{\bsnm{Huang}, \binits{S.}},
\bauthor{\bsnm{Wang}, \binits{X.}},
\bauthor{\bsnm{Tao}, \binits{D.}}:
\bctitle{Snapmix: Semantically proportional mixing for augmenting fine-grained
  data}.
In: \bbtitle{Proceedings of the AAAI Conference on Artificial Intelligence},
vol. \bseriesno{35},
pp. \bfpage{1628}--\blpage{1636}
(\byear{2021})
\end{bchapter}
\endbibitem

\bibitem{zhang2018adversarial}
\begin{bchapter}
\bauthor{\bsnm{Zhang}, \binits{X.}},
\bauthor{\bsnm{Wei}, \binits{Y.}},
\bauthor{\bsnm{Feng}, \binits{J.}},
\bauthor{\bsnm{Yang}, \binits{Y.}},
\bauthor{\bsnm{Huang}, \binits{T.S.}}:
\bctitle{Adversarial complementary learning for weakly supervised object
  localization}.
In: \bbtitle{Proceedings of the IEEE Conference on Computer Vision and Pattern
  Recognition},
pp. \bfpage{1325}--\blpage{1334}
(\byear{2018})
\end{bchapter}
\endbibitem

\bibitem{ding2021ap}
\begin{barticle}
\bauthor{\bsnm{Ding}, \binits{Y.}},
\bauthor{\bsnm{Ma}, \binits{Z.}},
\bauthor{\bsnm{Wen}, \binits{S.}},
\bauthor{\bsnm{Xie}, \binits{J.}},
\bauthor{\bsnm{Chang}, \binits{D.}},
\bauthor{\bsnm{Si}, \binits{Z.}},
\bauthor{\bsnm{Wu}, \binits{M.}},
\bauthor{\bsnm{Ling}, \binits{H.}}:
\batitle{Ap-cnn: Weakly supervised attention pyramid convolutional neural
  network for fine-grained visual classification}.
\bjtitle{IEEE Transactions on Image Processing}
\bvolume{30},
\bfpage{2826}--\blpage{2836}
(\byear{2021})
\end{barticle}
\endbibitem

\bibitem{dosovitskiy2020image}
\begin{botherref}
\oauthor{\bsnm{Dosovitskiy}, \binits{A.}},
\oauthor{\bsnm{Beyer}, \binits{L.}},
\oauthor{\bsnm{Kolesnikov}, \binits{A.}},
\oauthor{\bsnm{Weissenborn}, \binits{D.}},
\oauthor{\bsnm{Zhai}, \binits{X.}},
\oauthor{\bsnm{Unterthiner}, \binits{T.}},
\oauthor{\bsnm{Dehghani}, \binits{M.}},
\oauthor{\bsnm{Minderer}, \binits{M.}},
\oauthor{\bsnm{Heigold}, \binits{G.}},
\oauthor{\bsnm{Gelly}, \binits{S.}}, et al.:
An image is worth 16x16 words: Transformers for image recognition at scale.
arXiv preprint arXiv:2010.11929
(2020)
\end{botherref}
\endbibitem

\bibitem{hu2021rams}
\begin{bchapter}
\bauthor{\bsnm{Hu}, \binits{Y.}},
\bauthor{\bsnm{Jin}, \binits{X.}},
\bauthor{\bsnm{Zhang}, \binits{Y.}},
\bauthor{\bsnm{Hong}, \binits{H.}},
\bauthor{\bsnm{Zhang}, \binits{J.}},
\bauthor{\bsnm{He}, \binits{Y.}},
\bauthor{\bsnm{Xue}, \binits{H.}}:
\bctitle{Rams-trans: Recurrent attention multi-scale transformer for
  fine-grained image recognition}.
In: \bbtitle{Proceedings of the 29th ACM International Conference on
  Multimedia},
pp. \bfpage{4239}--\blpage{4248}
(\byear{2021})
\end{bchapter}
\endbibitem

\bibitem{zhang2022free}
\begin{bchapter}
\bauthor{\bsnm{Zhang}, \binits{Y.}},
\bauthor{\bsnm{Cao}, \binits{J.}},
\bauthor{\bsnm{Zhang}, \binits{L.}},
\bauthor{\bsnm{Liu}, \binits{X.}},
\bauthor{\bsnm{Wang}, \binits{Z.}},
\bauthor{\bsnm{Ling}, \binits{F.}},
\bauthor{\bsnm{Chen}, \binits{W.}}:
\bctitle{A free lunch from vit: adaptive attention multi-scale fusion
  transformer for fine-grained visual recognition}.
In: \bbtitle{ICASSP 2022-2022 IEEE International Conference on Acoustics,
  Speech and Signal Processing (ICASSP)},
pp. \bfpage{3234}--\blpage{3238}
(\byear{2022}).
\bcomment{IEEE}
\end{bchapter}
\endbibitem

\bibitem{rao2021counterfactual}
\begin{bchapter}
\bauthor{\bsnm{Rao}, \binits{Y.}},
\bauthor{\bsnm{Chen}, \binits{G.}},
\bauthor{\bsnm{Lu}, \binits{J.}},
\bauthor{\bsnm{Zhou}, \binits{J.}}:
\bctitle{Counterfactual attention learning for fine-grained visual
  categorization and re-identification}.
In: \bbtitle{Proceedings of the IEEE/CVF International Conference on Computer
  Vision},
pp. \bfpage{1025}--\blpage{1034}
(\byear{2021})
\end{bchapter}
\endbibitem

\bibitem{he2022transfg}
\begin{bchapter}
\bauthor{\bsnm{He}, \binits{J.}},
\bauthor{\bsnm{Chen}, \binits{J.-N.}},
\bauthor{\bsnm{Liu}, \binits{S.}},
\bauthor{\bsnm{Kortylewski}, \binits{A.}},
\bauthor{\bsnm{Yang}, \binits{C.}},
\bauthor{\bsnm{Bai}, \binits{Y.}},
\bauthor{\bsnm{Wang}, \binits{C.}}:
\bctitle{Transfg: A transformer architecture for fine-grained recognition}.
In: \bbtitle{Proceedings of the AAAI Conference on Artificial Intelligence},
vol. \bseriesno{36},
pp. \bfpage{852}--\blpage{860}
(\byear{2022})
\end{bchapter}
\endbibitem

\bibitem{sun2022sim}
\begin{bchapter}
\bauthor{\bsnm{Sun}, \binits{H.}},
\bauthor{\bsnm{He}, \binits{X.}},
\bauthor{\bsnm{Peng}, \binits{Y.}}:
\bctitle{Sim-trans: Structure information modeling transformer for fine-grained
  visual categorization}.
In: \bbtitle{Proceedings of the 30th ACM International Conference on
  Multimedia},
pp. \bfpage{5853}--\blpage{5861}
(\byear{2022})
\end{bchapter}
\endbibitem

\bibitem{miao2021complemental}
\begin{barticle}
\bauthor{\bsnm{Miao}, \binits{Z.}},
\bauthor{\bsnm{Zhao}, \binits{X.}},
\bauthor{\bsnm{Wang}, \binits{J.}},
\bauthor{\bsnm{Li}, \binits{Y.}},
\bauthor{\bsnm{Li}, \binits{H.}}:
\batitle{Complemental attention multi-feature fusion network for fine-grained
  classification}.
\bjtitle{IEEE Signal Processing Letters}
\bvolume{28},
\bfpage{1983}--\blpage{1987}
(\byear{2021})
\end{barticle}
\endbibitem

\bibitem{chou2022novel}
\begin{botherref}
\oauthor{\bsnm{Chou}, \binits{P.-Y.}},
\oauthor{\bsnm{Lin}, \binits{C.-H.}},
\oauthor{\bsnm{Kao}, \binits{W.-C.}}:
A novel plug-in module for fine-grained visual classification.
arXiv preprint arXiv:2202.03822
(2022)
\end{botherref}
\endbibitem

\bibitem{do2022fine}
\begin{botherref}
\oauthor{\bsnm{Do}, \binits{T.}},
\oauthor{\bsnm{Tran}, \binits{H.}},
\oauthor{\bsnm{Tjiputra}, \binits{E.}},
\oauthor{\bsnm{Tran}, \binits{Q.D.}},
\oauthor{\bsnm{Nguyen}, \binits{A.}}:
Fine-grained visual classification using self assessment classifier.
arXiv preprint arXiv:2205.10529
(2022)
\end{botherref}
\endbibitem

\bibitem{wang2021dynamic}
\begin{bchapter}
\bauthor{\bsnm{Wang}, \binits{S.}},
\bauthor{\bsnm{Li}, \binits{H.}},
\bauthor{\bsnm{Wang}, \binits{Z.}},
\bauthor{\bsnm{Ouyang}, \binits{W.}}:
\bctitle{Dynamic position-aware network for fine-grained image recognition}.
In: \bbtitle{Proceedings of the AAAI Conference on Artificial Intelligence},
vol. \bseriesno{35},
pp. \bfpage{2791}--\blpage{2799}
(\byear{2021})
\end{bchapter}
\endbibitem

\bibitem{vaswani2017attention}
\begin{botherref}
\oauthor{\bsnm{Vaswani}, \binits{A.}},
\oauthor{\bsnm{Shazeer}, \binits{N.}},
\oauthor{\bsnm{Parmar}, \binits{N.}},
\oauthor{\bsnm{Uszkoreit}, \binits{J.}},
\oauthor{\bsnm{Jones}, \binits{L.}},
\oauthor{\bsnm{Gomez}, \binits{A.N.}},
\oauthor{\bsnm{Kaiser}, \binits{{\L}.}},
\oauthor{\bsnm{Polosukhin}, \binits{I.}}:
Attention is all you need.
Advances in neural information processing systems
\textbf{30}
(2017)
\end{botherref}
\endbibitem

\bibitem{shazeer2020talking}
\begin{botherref}
\oauthor{\bsnm{Shazeer}, \binits{N.}},
\oauthor{\bsnm{Lan}, \binits{Z.}},
\oauthor{\bsnm{Cheng}, \binits{Y.}},
\oauthor{\bsnm{Ding}, \binits{N.}},
\oauthor{\bsnm{Hou}, \binits{L.}}:
Talking-heads attention.
arXiv preprint arXiv:2003.02436
(2020)
\end{botherref}
\endbibitem

\bibitem{kipf2016semi}
\begin{botherref}
\oauthor{\bsnm{Kipf}, \binits{T.N.}},
\oauthor{\bsnm{Welling}, \binits{M.}}:
Semi-supervised classification with graph convolutional networks.
arXiv preprint arXiv:1609.02907
(2016)
\end{botherref}
\endbibitem

\bibitem{zhou2020look}
\begin{bchapter}
\bauthor{\bsnm{Zhou}, \binits{M.}},
\bauthor{\bsnm{Bai}, \binits{Y.}},
\bauthor{\bsnm{Zhang}, \binits{W.}},
\bauthor{\bsnm{Zhao}, \binits{T.}},
\bauthor{\bsnm{Mei}, \binits{T.}}:
\bctitle{Look-into-object: Self-supervised structure modeling for object
  recognition}.
In: \bbtitle{Proceedings of the IEEE/CVF Conference on Computer Vision and
  Pattern Recognition},
pp. \bfpage{11774}--\blpage{11783}
(\byear{2020})
\end{bchapter}
\endbibitem

\bibitem{luo2020learning}
\begin{barticle}
\bauthor{\bsnm{Luo}, \binits{W.}},
\bauthor{\bsnm{Zhang}, \binits{H.}},
\bauthor{\bsnm{Li}, \binits{J.}},
\bauthor{\bsnm{Wei}, \binits{X.-S.}}:
\batitle{Learning semantically enhanced feature for fine-grained image
  classification}.
\bjtitle{IEEE Signal Processing Letters}
\bvolume{27},
\bfpage{1545}--\blpage{1549}
(\byear{2020})
\end{barticle}
\endbibitem

\bibitem{huang2021stochastic}
\begin{bchapter}
\bauthor{\bsnm{Huang}, \binits{S.}},
\bauthor{\bsnm{Wang}, \binits{X.}},
\bauthor{\bsnm{Tao}, \binits{D.}}:
\bctitle{Stochastic partial swap: Enhanced model generalization and
  interpretability for fine-grained recognition}.
In: \bbtitle{Proceedings of the IEEE/CVF International Conference on Computer
  Vision},
pp. \bfpage{620}--\blpage{629}
(\byear{2021})
\end{bchapter}
\endbibitem

\bibitem{zhu2022dual}
\begin{bchapter}
\bauthor{\bsnm{Zhu}, \binits{H.}},
\bauthor{\bsnm{Ke}, \binits{W.}},
\bauthor{\bsnm{Li}, \binits{D.}},
\bauthor{\bsnm{Liu}, \binits{J.}},
\bauthor{\bsnm{Tian}, \binits{L.}},
\bauthor{\bsnm{Shan}, \binits{Y.}}:
\bctitle{Dual cross-attention learning for fine-grained visual categorization
  and object re-identification}.
In: \bbtitle{Proceedings of the IEEE/CVF Conference on Computer Vision and
  Pattern Recognition},
pp. \bfpage{4692}--\blpage{4702}
(\byear{2022})
\end{bchapter}
\endbibitem

\bibitem{wang2021feature}
\begin{botherref}
\oauthor{\bsnm{Wang}, \binits{J.}},
\oauthor{\bsnm{Yu}, \binits{X.}},
\oauthor{\bsnm{Gao}, \binits{Y.}}:
Feature fusion vision transformer for fine-grained visual categorization.
arXiv preprint arXiv:2107.02341
(2021)
\end{botherref}
\endbibitem

\bibitem{bera2022sr}
\begin{barticle}
\bauthor{\bsnm{Bera}, \binits{A.}},
\bauthor{\bsnm{Wharton}, \binits{Z.}},
\bauthor{\bsnm{Liu}, \binits{Y.}},
\bauthor{\bsnm{Bessis}, \binits{N.}},
\bauthor{\bsnm{Behera}, \binits{A.}}:
\batitle{Sr-gnn: Spatial relation-aware graph neural network for fine-grained
  image categorization}.
\bjtitle{IEEE Transactions on Image Processing}
\bvolume{31},
\bfpage{6017}--\blpage{6031}
(\byear{2022})
\end{barticle}
\endbibitem

\end{thebibliography}


\end{document}